\documentclass[letterpaper]{article} 
\usepackage[]{aaai25}  
\usepackage{times}  
\usepackage{helvet}  
\usepackage{courier}  
\usepackage[hyphens]{url}  
\usepackage{graphicx} 
\usepackage{natbib}  
\usepackage{caption} 
\usepackage{algorithm}
\usepackage{algorithmic}

\usepackage{colortbl}
\usepackage{xcolor}
\usepackage{rotating}
\usepackage{color}
\usepackage{subfigure}
\usepackage{multirow}
\usepackage{amsmath}
\usepackage{amssymb}
\usepackage{booktabs}
\usepackage{pifont}
\usepackage{makecell}
\usepackage{fancyvrb}

\usepackage{xcolor}

\usepackage{newfloat}
\usepackage{listings}
\DeclareCaptionStyle{ruled}{labelfont=normalfont,labelsep=colon,strut=off} 
\floatstyle{ruled}
\newfloat{listing}{tb}{lst}{}
\floatname{listing}{Listing}
\title {A Comprehensive Evaluation of Large Language Models on Aspect-Based Sentiment Analysis}
\author {
    Changzhi Zhou\textsuperscript{\rm 1},
    Dandan Song\textsuperscript{\rm 1}\thanks{~ Corresponding Author},
    Yuhang Tian\textsuperscript{\rm 1},
    Zhijing Wu\textsuperscript{\rm 1},
    Hao Wang\textsuperscript{\rm 1},
    Xinyu Zhang\textsuperscript{\rm 1},
    Jun Yang\textsuperscript{\rm 1},
    Ziyi Yang
    \textsuperscript{\rm 1},
    Shuhao Zhang
    \textsuperscript{\rm 2}
}
\affiliations {
    \textsuperscript{\rm 1}School of Computer Science and Technology, Beijing Institute of Technology, Beijing, China\\
    \textsuperscript{\rm 2}Nanyang Technological University, Singapore\\
    zhou\_changzhi97@bit.edu.cn
}

\usepackage{bibentry}

\begin{document}
\maketitle
\begin{abstract}

Recently, Large Language Models (LLMs) have garnered increasing attention in the field of natural language processing, revolutionizing numerous downstream tasks with powerful reasoning and generation abilities. For example, In-Context Learning (ICL) introduces a fine-tuning-free paradigm, allowing out-of-the-box LLMs to execute downstream tasks by analogy learning without any fine-tuning. Besides, in a fine-tuning-dependent paradigm where substantial training data exists, Parameter-Efficient Fine-Tuning (PEFT), as the cost-effective methods, enable LLMs to achieve excellent performance comparable to full fine-tuning.

However, these fascinating techniques employed by LLMs have not been fully exploited in the ABSA field. Previous works probe LLMs in ABSA by merely using randomly selected input-output pairs as demonstrations in ICL, resulting in an incomplete and superficial evaluation. In this paper, we shed light on a comprehensive evaluation of LLMs in the ABSA field, involving 13 datasets, 8 ABSA subtasks, and 6 LLMs. Specifically, we design a unified task formulation to unify ``multiple LLMs for multiple ABSA subtasks in multiple paradigms.'' For the fine-tuning-dependent paradigm, we efficiently fine-tune LLMs using instruction-based multi-task learning. For the fine-tuning-free paradigm, we propose 3 demonstration selection strategies to stimulate the few-shot abilities of LLMs. Our extensive experiments demonstrate that LLMs achieve a new state-of-the-art performance compared to fine-tuned Small~\footnote{\textit{Small} is a relative wording. In this paper, we define models with $<$1B parameters are small (e.g., BART-base). Conversely, they are large.} Language Models (SLMs) in the fine-tuning-dependent paradigm. More importantly, in the fine-tuning-free paradigm where SLMs are ineffective, LLMs with ICL still showcase impressive potential and even compete with fine-tuned SLMs on some ABSA subtasks.

\end{abstract}

\section{Introduction}

\begin{figure}[t]
	\centering
	\includegraphics[width=1.0\columnwidth]{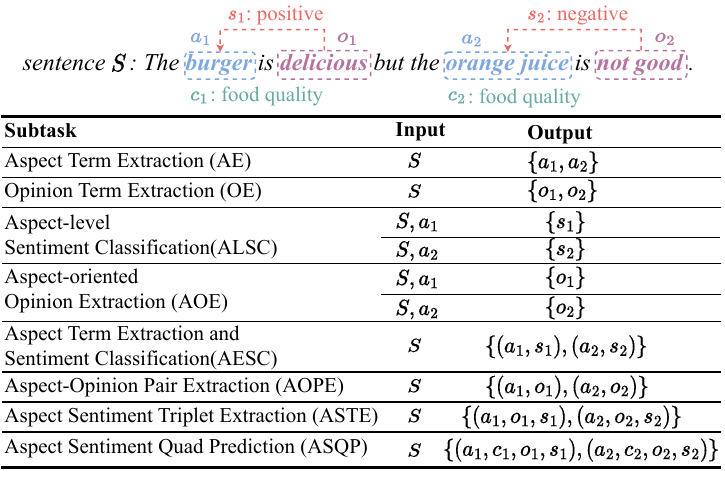}
	\caption{An illustration of different ABSA subtasks. The $a, c, o$, and $s$ denote \textit{aspect term}, \textit{aspect category}, \textit{opinion term}, and \textit{sentiment polarity}, respectively.
	}
	\label{intro:case}
\end{figure}

Aspect-Based Sentiment Analysis (ABSA) is the fine-grained Sentiment Analysis (SA) task that attracts widespread attention from academia and industry due to its practical applications~\cite{liu2012sentiment, pontiki-etal-2016-semeval}. Typically, ABSA involves four sentiment elements: \textit{aspect term}, \textit{aspect category}, \textit{opinion term} and \textit{sentiment polarity}~\cite{zhang2023survey}. For example, in the sentence shown in Figure~\ref{intro:case}, the four corresponding sentiment elements are \textit{\{burger, orange juice\}}, \textit{\{food quality, food quality\}}, \textit{\{delicious, not good\}}, and \textit{\{positive, negative\}}. Based on the types of these elements in the input and output, ABSA can be divided into different subtasks.

Many previous studies tailor complicated models for a single subtask. For example, \citet{fan-etal-2019-target} propose a target-fused neural sequence labeling method for the AOE subtask, and \citet{Peng_Xu_Bing_Huang_Lu_Si_2020} propose a two-stage framework, first extracting elements and then identifying relations, for the ASTE subtask. However, this type of work struggles to generalize to other subtasks. Subsequently, some works fine-tune sequence-to-sequence models to unify multiple subtasks. BARTABSA~\cite{yan-etal-2021-unified} pioneers using a BART-base model~\cite{lewis-etal-2020-bart} to accomplish seven subtasks by reframing each subtask target as the generation of indexes. Paraphrase~\cite{zhang-etal-2021-aspect-sentiment} based on the T5-base model~\cite{t5} unifies multiple subtasks by linearizing sentiment tuples to a natural language sequence. These crafted Small Language Models (SLMs) obtain state-of-the-art performance when fine-tuned with ample data. However, many real-world applications involve low-resource scenarios, such as narrow domains or complex tasks, where data annotation is time-consuming and labor-intensive. SLMs often ineffective in these data-scarce settings~\cite{zhang2023survey}.

Recently, Large Language Models (LLMs) have demonstrated powerful abilities due to their large number of parameters and extensive pre-training data~\cite{zhao2023survey}. On the one hand, In-Context Learning (ICL), one of the emergent abilities of LLMs, pioneers a new fine-tuning-free paradigm~\cite{brown2020language,dong2023survey}. By providing demonstrations consisting of input-output examples, LLMs can learn by analogy like humans without any fine-tuning. When faced with fine-tuning-dependent paradigm with sufficient data, on the other hand, Parameter-Efficient Fine-Tuning (PEFT) methods~\cite{han2024peftsurvey} allow LLMs to obtain powerful performance on downstream tasks in a cost-effective manner. With these innovative technologies, LLMs have revolutionized many tasks in the field of natural language processing~\cite{zhao2023survey}. However, the majority of existing ABSA research still focuses on SLMs~\cite{zhang2023survey}. Some recent works~\cite{zhang2023sentiment, amin2023wide,wang2024chatgpt} conduct preliminary evaluation of LLMs for ABSA, but they are not comprehensive or in-depth. These evaluations either focus on a single ABSA subtask, a single LLM, or a single demonstration selection strategy in ICL.

In order to achieve a comprehensive evaluation of LLMs for ABSA subtasks, in this paper, we conduct extensive experiments on 13 datasets of 8 ABSA subtasks by fine-tuning open-source LLMs (fine-tuned LLMs) or calling out-of-the-box LLMs (API-based LLMs). The former aims to explore whether LLMs can surpass SLMs in the fine-tuning-dependent paradigm. The latter studies the zero-/few-shot abilities of LLMs and the effectiveness of demonstration selection strategies of ICL in the fine-tuning-free paradigm, where SLMs are ineffective. Concretely, we integrate existing ABSA datasets and propose an unified task formulation to unify ``multiple LLMs for multiple ABSA subtasks in multiple paradigms''. We use instruction-based multi-task learning and low-rank adaptation (LoRA), a computationally cost-friendly method with performance comparable to full fine-tuning~\cite{hu2022lora}, to efficiently fine-tune open-source LLMs. Besides, we construct three demonstration selection strategies, \textit{random-based selection}, \textit{keyword-based selection} and \textit{semantic-based selection}, to unlock the potential of ICL. 








Our experimental results and analysis yield the following insightful conclusions: \textbf{Firstly}, in the fine-tuning-dependent paradigm, LLMs outperform full fine-tuned SLMs on all ABSA subtasks even with minimal parameter fine-tuning (e.g. LLaMA3-8B with 3.4M vs. T5-base with 220M). \textbf{More importantly}, in the fine-tuning-free paradigm where SLMs fail completely, API-based LLMs with ICL achieve remarkable performance, even matching that of fine-tuned SLMs in some subtasks. This sheds light on future ABSA research in low-resource scenarios. \textbf{Besides}, the \textit{keyword} and \textit{semantic information} in ICL can significantly improve the performance of LLMs, and combining these two orthogonal types of information can lead to better effectiveness. \textbf{Finally}, we have novel findings that the effectiveness of ICL varies depending on the specific ABSA subtask and LLMs used. In some cases, ICL can even cause adverse effects. More conclusions and analysis can be found in the Experiments section. The contributions of this paper are as follows:

1) We thoroughly evaluate the performance of LLMs in the ABSA field, involving 13 datasets, 8 subtasks, and 6 LLMs, achieving ``multiple LLMs for multiple ABSA subtasks in multiple paradigms with an unified formulation.''

2) We demonstrate that efficiently fine-tuned LLMs using instruction-based multi-task learning can comprehensively outperform fine-tuned SLMs in the fine-tuning-dependent paradigm.

3) We study different demonstration selection strategies and greatly improve the performance of API-based LLMs in the fine-tuning-free paradigm, where SLMs fail completely.



In summary, our work demonstrates that, regardless of the paradigm, LLMs always outperform SLMs. We hope that the comprehensive LLM-based baselines established by this paper will promote the development of LLMs in the ABSA field.




\section{Related Work}

\subsection{Aspect-Based Sentiment Analysis}

	Aspect-Based Sentiment Analysis (ABSA), a fine-grained sentiment analysis problem, attracts considerable attention owing to its practicability~\cite{liu2012sentiment, pontiki-etal-2014-semeval, zhang2023survey}. Previous works~\cite{Peng_Xu_Bing_Huang_Lu_Si_2020, Wan_Yang_Du_Liu_Qi_Pan_2020, cai-etal-2021-aspect,zhang-etal-2022-boundary, zhou2024diaasq} aim to develop specialized models tailored for separate ABSA subtasks, which limits their scope of application. Therefore, more recent works attempt to unify multiple ABSA subtasks, namely ``one model for all subtasks''~\cite{wang2024unifiedabsa}. Concretely, BARTABSA~\cite{yan-etal-2021-unified} employs the BART model and the pointer mechanism~\cite{NIPS2015_29921001} to convert extraction and classification tasks into generating pointer indexes and class indexes. \citet{Mao_Shen_Yu_Cai_2021} develop a BERT-based machine reading comprehension framework with a dual structure (Dual-MRC). \citet{zhang-etal-2021-towards-generative, zhang-etal-2021-aspect-sentiment, mao-etal-2022-seq2path} formulate ABSA subtasks as a text-generation problem to utilize label information and the power of the T5 model. Furthermore, LEGO-ABSA~\cite{gao-etal-2022-lego} propose a T5-based unified generative framework that solves multiple subtasks simultaneously by multi-task learning. MVP~\cite{gou-etal-2023-mvp} provides element order prompts to direct the T5 model in generating numerous tuples, each with a distinct element order. UnifiedABSA~\cite{wang2024unifiedabsa} decouples the quadruple labels and designs task-specific instructions.

    \subsection{Large Language Models for ABSA}
    
    The development of Large Language Models (LLMs) has revolutionized the field of natural language processing, such as Retrieval-Augmented Generation~\cite{gao2024retrievalaugmented} and AI Agents~\cite{guo2024large}. LLMs have also received extensive attention in traditional information extraction tasks such as Named Entity Recognition and Relation Extraction~\cite{wadhwa-etal-2023-revisiting, li-etal-2023-revisiting-large, wang2023gptner, xu2024large}. However, research on LLMs for ABSA tasks is still nascent. Some works~\cite{zhang2023sentiment, amin2023wide, zhao2023chatgpt, wang2024chatgpt} preliminarily evaluate the performance of LLMs for multiple SA tasks in los-resource scenarios, but the focus on ABSA has been relatively minimal. Besides, \citet{simmering2023large} achieve state-of-the-art performance on the AESC task by designing sophisticated prompts and fine-tuning GPT-3.5. \citet{xu2023limits} explore the potential of ICL for the ASQP subtask. However, they all ignore other ABSA subtasks. A comprehensive exploration of LLMs for multiple ABSA subtasks remains a gap.

\begin{figure}[t]
		\centering
\includegraphics[width=1.0\columnwidth]{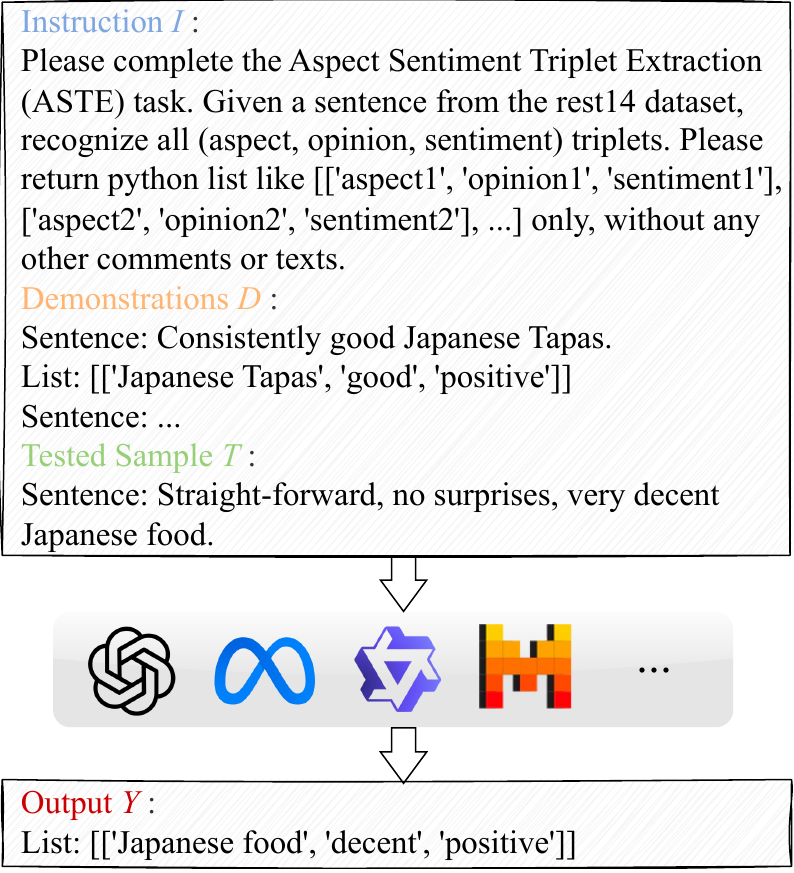}
		\caption{An example of the ASTE task formulation. 
		}
		\label{method:model}
\end{figure}

\section{Methodology}
\label{sec:three}

\subsection{Unified Task Formulation}
\label{sec:three.one}

We unify various subtasks into a list generation pattern by designing various prompt and output for LLMs. This pattern can be used for API-based LLMs and find-tuned LLMs simultaneously. As shown in Figure~\ref{method:model}, the prompt ($X$) includes: 

\textbf{Instruction} ($I$): We design unique instructions for each ABSA subtask, guiding LLMs on what subtask to complete and which sentiment element to return.

\textbf{Demonstrations} ($D$): Several input-output pairs compose demonstrations, from which LLMs can learn input-output mapping rules. Unless otherwise indicated, demonstrations are enabled solely in the fine-tuning-free paradigm.


\textbf{Tested sample} ($T$): It includes a sentence and a aspect term in ALSC and AOE subtasks and only a sentence in other subtasks. 


The \textbf{output} ($Y$) is two-dimensional list of sentiment elements corresponding to the particular subtask. 





\begin{table}[t]
  \centering
  \small
    \begin{tabular}{llcccc}
    \toprule
    \multicolumn{2}{c}{} & training & validation & test  & subtasks \\
    \midrule
    \multirow{3}[2]{*}{$\mathcal{D}_{17}$} & L14   & 3048  & /     & 800   & \multirow{2}[2]{*}{\textit{AE,OE,}} \\
          & R14   & 3044  & /     & 800   & \multirow{2}[2]{*}{\textit{ALSC}} \\
          & R15   & 1315  & /     & 685   & \\
    \midrule
    \multirow{4}[2]{*}{$\mathcal{D}_{19}$} & L14   & 1158  & /     & 343   & \multirow{4}[2]{*}{\textit{AOE}} \\
          & R14   & 1627  & /     & 500   &  \\
          & R15   & 754   & /     & 325   &  \\
          & R16   & 1079  & /     & 329   &  \\
    \midrule
    \multirow{4}[2]{*}{$\mathcal{D}_{20}$} & L14   & 920   & 228   & 339   & \multirow{2}[2]{*}{\textit{AESC,}} \\
          & R14   & 1300  & 323   & 496   & \multirow{2}[2]{*}{\textit{AOPE,}} \\
          & R15   & 593   & 148   & 318   & \multirow{2}[2]{*}{\textit{ASTE} } \\
          & R16   & 842   & 210   & 320   & \\
    \midrule
    \multirow{2}[2]{*}{$\mathcal{D}_{21}$} & R15   & 834   & 209   & 537   & \multirow{2}[2]{*}{\textit{ASQP}} \\
          & R16   & 1264  & 316   & 544   &  \\
    \bottomrule
    \end{tabular}%
    \caption{Statistics of ABSA datasets. $\mathcal{D}_{17}$, $\mathcal{D}_{19}$,  $\mathcal{D}_{20}$, and $\mathcal{D}_{21}$ contains three, four, four, and two datasets, respectively. L and R denotes laptop and restaurant, respectively.}
  \label{exp:dataset}%
\end{table}%

\subsection{Demonstration Selection Strategy}
\label{sec:icl}
To explore the differences between different demonstrations and stimulate the few-shot abilities of LLMs, we utilize three selection strategies as follows:

\textbf{Random-based selection}: It is a vanilla demonstration selection strategy. 

\textbf{Keyword-based selection}: BM25~\cite{2009bm25} is a sparse retrieval function. It calculates the relevance score between sentences based on term frequency (TF), inverse document frequency (IDF) and sentence length. Therefore, the sentence pairs retrieved by BM25 share common keywords, which may be aspect terms or opinion terms to be extracted.

\textbf{Semantic-based selection}: SimCSE~\cite{gao-etal-2021-simcse} is a popular semantic-based dense retrieval model, which obtains sentence embeddings by leveraging contrastive learning. We use it to select semantically similar demonstrations.

\section{Experiments}

\subsection{Experimental Setup}

\subsubsection{Datasets and Metrics}

\setlength{\tabcolsep}{4pt} 
\begin{table*}[]
  \centering
    \begin{tabular}{lccc|ccc|ccc|cccc|c}
    \toprule[1.2pt]
    \multicolumn{1}{l}{\multirow{2}[4]{*}{Methods}}  & \multicolumn{3}{c}{AE} & \multicolumn{3}{c}{OE} & \multicolumn{3}{c}{ALSC} & \multicolumn{4}{c}{AOE} & 
 \multirow{2}[2]{*}{\textbf{AVG}} \\
\cmidrule{2-14}           & L14   & R14   & R15   & L14   & R14   & R15   & L14   & R14   & R15   &  L14   & R14   & R15  & R16 & \\
    \midrule
    \rowcolor{gray!15}\multicolumn{15}{l}{\textbf{Full Fine-tuned SLMs with full dataset}} \\
    Dual-MRC &  82.51  & 86.60  & 75.08  & /     & /   & /  & 75.97  & 82.04  & 73.59   & 79.90 &	83.73 &	74.50 &	83.33 & \cellcolor{gray!15}/
 \\
    DCRAN &   \textbf{85.61}  & \textbf{89.67} & \textbf{79.68}  & 79.77  & \textbf{87.59}  & \textbf{79.90}  & \textbf{80.78}  & \textbf{84.22}  & 77.99  &  / & / & /  & /  & \cellcolor{gray!15}/  \\
    BARTABSA &  83.52  & 87.07  & 75.48  & 77.86  & 87.29  & 76.49  & 76.76  & 75.56  & 73.91  & 80.55 &	85.38 &	80.52 &	87.92 & \cellcolor{gray!15}80.64  \\
    T5-Instruct  & 84.05  & 87.51  & 77.22  & \textbf{80.91}  & 86.55  & 78.64  & 78.29  & 82.82  & \textbf{86.27}  & \textbf{81.04}  & \textbf{85.55}  & \textbf{84.92}  & \textbf{89.12}  & \cellcolor{gray!15}\textbf{83.30}  \\
    \rowcolor{gray!15}\multicolumn{15}{l}{\textbf{Efficiently Fine-tuned LLMs with full dataset}} \\
    ChatGLM3-6B  & 83.28  & 87.28  & 76.79  & 82.80  & 85.49  & 77.42  & 81.04  & 87.35  & 87.27  & 81.79  & 87.29  & 82.57  & 88.08  & \cellcolor{gray!15}83.73  \\
    QWen1.5-7B  & \textbf{87.87}  & 87.95  & \textbf{80.54}  & 81.48  & 87.26  & 79.29  & 81.19 &	88.14 &	88.75   & 83.73  & 88.14  & 84.84  & \textbf{91.67}  & \cellcolor{gray!15}85.68  \\
Mistral-7B-v0.2 & 85.53  & 88.87  & 79.66  & 82.26  & \textbf{87.60}  & 79.59  & \textbf{81.50}  & 88.27  & 88.75  & \textbf{84.72}  & 87.86  & 85.63  & 91.18  & \cellcolor{gray!15}85.49  \\
    
    LLaMA3-8B & 87.62  & \textbf{88.97}  & 79.67  & \textbf{83.33} & 87.40 & \textbf{82.15} & 80.73  & \textbf{88.63}  & \textbf{89.11}  & 84.11   & 	\textbf{88.90}   & 	\textbf{85.71}   & 	91.34 & \cellcolor{gray!15}\textbf{86.17} 
 \\
 \midrule[1pt]
    \rowcolor{gray!15}\multicolumn{15}{l}{\textbf{API-based LLMs without fine-tuning (zero-/three-shot)}} \\
    LLaMA3-70B	&	 53.96  & 72.28  & 56.11  & 65.88  & 74.35  & 58.34  & 77.37  & 84.12  & 85.71  & 64.52  & 74.43  & 67.43  & 76.16  & \cellcolor{gray!15}70.05  \\
    \ \ \textit{+Random} 	 & 54.33  & 73.71  &  \underline{55.94}  & 66.44  & 77.44  &  \underline{56.82}  & 79.20  & \textbf{85.58}  & \textbf{86.16}  &  \underline{63.76}  & 75.11  & 69.53  & 78.03  & \cellcolor{gray!15}70.93  \\
    \ \ \textit{+BM25} 	 & 59.87  & 78.03  & 64.66  & \textbf{68.55}  & \textbf{80.56}  & \textbf{62.24}  & 78.13  & 85.22  & 85.79  & 67.29  &  \underline{74.28}  & 68.39  & 77.32  & \cellcolor{gray!15}73.10  \\
    \ \ \textit{+SimCSE} 	 & 	60.41  & 78.88  & 63.99  & 66.71  & 73.59  & 59.00  & 78.75  & 84.78  & 85.79  & 68.10  & 74.63  & 69.05  & 78.14  & \cellcolor{gray!15}72.45  \\
    \cmidrule{2-15}
    ChatGPT	&	  50.91  & 68.38  & 50.92  & 55.89  & 71.34  & 51.83  & 74.48  & 81.23  & 82.56  & 59.91  & 67.73  & 66.15  & 70.50 & \cellcolor{gray!15}65.53 \\
    \ \ \textit{+Random}  & 51.37  & 70.53  & 54.06  & 61.05  & 74.18  & 56.10  & 76.14  & 83.47  & 84.06  & 62.91  & 72.09  & 67.77  & 74.52 & \cellcolor{gray!15}68.33 \\
    \ \ \textit{+BM25}  & 54.60  & 72.31  & 55.98  & 61.13  & 75.35  & 55.02  & 76.30  & 81.87  & 83.81  & 63.29  & 70.38  & 68.17  & 73.64 & \cellcolor{gray!15}68.60 \\
    \ \ \textit{+SimCSE}  & 54.90  & 72.10  & 57.77  & 59.56  & 74.95  & 55.19  & 75.33  & 81.76  & 82.81  & 64.74  & 71.88  & 69.33  & 77.08 & \cellcolor{gray!15}69.03 \\
    \cmidrule{2-15}
    GPT4+\textit{BM25} & \textbf{66.11}  & \textbf{80.67}  & \textbf{69.60}  & 66.55  & 79.43  & 61.20  & \textbf{79.66}  & 84.90  & 85.50  & \textbf{73.12}  & \textbf{79.62}  & \textbf{76.46}  & \textbf{83.05}  & \cellcolor{gray!15}\textbf{75.84}  \\
    \bottomrule[1.2pt]
    \end{tabular}%
    \caption{Main results of AE, OE, ALSC, and AOE subtasks. The AVG is the abbreviation of ``average''. \textbf{Bold} indicates state-of-the-art performance of a class of methods on a particular dataset.  \underline{Underline} indicates that the performance decreases after using demonstrations. The same notation definitions apply to Tables~\ref{exp:main2} and~\ref{exp:main3}.}
  \label{exp:main1}%
\end{table*}%

\setlength{\tabcolsep}{5pt} 
\begin{table*}[]
  \centering
    \begin{tabular}{lcccc|cccc|cccc|c}
    \toprule[1.2pt]
    \multicolumn{1}{l}{\multirow{2}[4]{*}{Methods}}  & \multicolumn{4}{c}{AESC} & \multicolumn{4}{c}{AOPE} & \multicolumn{4}{c}{ASTE} & \multirow{2}[4]{*}{\textbf{AVG}} \\
\cmidrule{2-13}   & L14   & R14   & R15   & \multicolumn{1}{c|}{R16}   & L14   & R14   & R15   & \multicolumn{1}{c|}{R16}   & L14   & R14   & R15   & R16   &  \\
    \midrule
    \rowcolor{gray!15}\multicolumn{14}{l}{\textbf{Full Fine-tuned SLMs with full dataset}} \\
    \multicolumn{1}{l}{TAGS}  &  /     & /     & /     & /     & 71.20  & \textbf{78.61}  & 72.23  & \textbf{80.47}  & \textbf{64.53}  & \textbf{75.05}  & \textbf{67.90}  & \textbf{76.61}  & \cellcolor{gray!15}/ \\
    \multicolumn{1}{l}{BARTABSA}  &  68.17  & 78.47  & 69.95  & 75.69  & 66.11  & 77.68  & 67.98  & 77.38  & 57.59  & 72.46  & 60.11  & 69.98  & \cellcolor{gray!15}70.13  \\
    \multicolumn{1}{l}{LEGO-ABSA}  &  72.30  & 80.60  & 74.20  & 76.10  & \textbf{71.30}  & 78.00  & \textbf{72.90}  & 77.10  & 62.20  & 73.70  & 64.40  & 69.90  & \cellcolor{gray!15}72.73  \\
    \multicolumn{1}{l}{T5-Instruct}  & \textbf{69.66}  & \textbf{83.43}  & \textbf{78.56}  & \textbf{79.31}  & 69.17  & 76.78  & 72.27  & 79.22  & 60.86  & 72.82  & 64.86  & 72.57  & \cellcolor{gray!15}\textbf{73.29}  \\
    \rowcolor{gray!15}\multicolumn{14}{l}{\textbf{Efficiently Fine-tuned LLMs with full dataset}} \\
    ChatGLM3-6B  & 71.96  & 83.01  & 77.02  & 79.80  & 68.80  & 79.14  & 74.67  & 75.41  & 62.05  & 74.65  & 67.11  & 70.30  & \cellcolor{gray!15}73.66  \\
    QWen1.5-7B  & \textbf{75.66}  & 83.36  & \textbf{81.65}  & 82.57  & 73.63  & 80.71  & 76.03  & \textbf{82.54}  & \textbf{67.36}  & 76.98  & 71.14  & \textbf{77.45}  & \cellcolor{gray!15}77.42  \\
    Mistral-7B-v0.2  & 73.12  & \textbf{84.49}  & 79.40  & 81.80  & 71.49  & 79.74  & 75.41  & 80.60  & 63.83  & 76.44  & 71.07  & 76.99  & \cellcolor{gray!15}76.20  \\
    LLaMA3-8B  &  74.52  & 84.20  & 80.07  & \textbf{84.13}  & \textbf{73.76}  & \textbf{81.95}  & \textbf{80.04}  & 81.74  & 65.68  & \textbf{79.20}  & \textbf{74.06}  & 76.15  & \cellcolor{gray!15}\textbf{77.96}  \\
    \midrule[1pt]
    \rowcolor{gray!15}\multicolumn{14}{l}{\textbf{API-based LLMs without fine-tuning (zero-/three-shot)}} \\
    LLaMA3-70B	 &  51.93  & 68.34  & 63.19  & 62.87  & 46.05  & 64.35  & 58.28  & 61.88  & 23.93  & 48.22  & 39.28  & 48.69  & \cellcolor{gray!15}53.08  \\
    \ \ \textit{+Random}  &  56.08  & 71.68  & 67.72  & 68.13  & 48.06  &  \underline{57.96}  &  \underline{54.86}  & 65.14  & 41.52  & 60.13  & 52.13  & 54.98  & \cellcolor{gray!15}58.20  \\
    \ \ \textit{+BM25}  &  59.53  & 72.03  & 70.39  & 69.89  & 51.94  & 67.17  & 65.54  & 68.00  & 43.99  & 56.84  & 54.93  & 58.58  & \cellcolor{gray!15}61.57  \\
    \ \ \textit{+SimCSE}  &  62.00  & 72.48  & 70.62  & 71.17  & 52.55  & 65.25  & 63.72  & 67.40  & 42.73  & 59.76  & 56.59  & 60.05  & \cellcolor{gray!15}62.03  \\
    \cmidrule{2-14}
    \multicolumn{1}{l}{ChatGPT} &  45.07  & 59.10  & 51.42  & 56.63  & 36.88  & 54.93  & 46.24  & 53.89  & 31.56  & 49.60  & 41.73  & 49.14  & \cellcolor{gray!15}48.02  \\
    \multicolumn{1}{l}{\ \ \textit{+Random}} &  49.49  & 65.72  & 59.42  & 64.07  & 43.13  & 56.00  & 51.71  & 56.81  & 35.71  & 55.10  & 46.92  & 52.71  & \cellcolor{gray!15}53.07  \\
    \multicolumn{1}{l}{\ \ \textit{+BM25}} &  54.48  & 67.48  & 64.35  & 63.31  & 43.79  & 60.48  & 57.77  & 58.72  & 42.71  & 58.04  & 49.91  & 54.69  & \cellcolor{gray!15}56.31  \\
    \multicolumn{1}{l}{\ \ \textit{+SimCSE}} &  56.53  & 67.92  & 60.79  & 63.02  & 46.65  & 58.42  & 55.81  & 58.57  & 41.73  & 55.44  & 49.67  & 54.75  & \cellcolor{gray!15}55.78  \\
    \cmidrule{2-14}
    GPT4+\textit{BM25}  &  \textbf{63.63} & \textbf{75.45} & \textbf{74.07} & \textbf{74.43} & \textbf{55.34} & \textbf{68.29} & \textbf{66.87} & \textbf{69.39} & \textbf{50.81} & \textbf{64.44} & \textbf{59.34} & \textbf{63.29} & \cellcolor{gray!15}\textbf{65.45} \\
    \bottomrule[1.2pt]
    \end{tabular}%
     \caption{Main results of AESC, AOPE, and ASTE subtasks.}
  \label{exp:main2}%
\end{table*}%

We evaluate all models on 13 datasets over 8 ABSA subtasks. All of the datasets originate from the Semeval Challenges~\cite{pontiki-etal-2014-semeval, pontiki-etal-2015-semeval, pontiki-etal-2016-semeval}. Detailed statistics of datasets are shown in Table~\ref{exp:dataset}. Following~\citet{yan-etal-2021-unified}, we leverage $\mathcal{D}_{17}$ annotated by~\citet{Wang_Pan_Dahlmeier_Xiao_2017} for AE, OE and ALSC subtasks, $\mathcal{D}_{19}$ annotated by~\citet{fan-etal-2019-target} for AOE subtask, $\mathcal{D}_{20}$ annotated by ~\citet{Peng_Xu_Bing_Huang_Lu_Si_2020} for AESC, AOPE and ASTE subtasks, and $\mathcal{D}_{21}$ of the ASQP subtask come from~\citet{zhang-etal-2021-aspect-sentiment}. Besides, to efficiently fine-tune LLMs using instruction-based multi-task learning, we combine all datasets as in previous work~\cite{gou-etal-2023-mvp}. Concretely, we combine all training sets and validation sets, and delete samples that overlap with any test sets. Then we split the data by 9:1 to obtain the final training and validation sets. In order to ensure a fair comparison with baselines, we use the F1 score (\%) as the evaluation metric.

\setlength{\tabcolsep}{4pt}
\begin{table}[]
  \centering
    \begin{tabular}{lccccc}
    \toprule[1.2pt]
    Methods & ASQP-R15 & ASQP-R16 & AVG \\
    \midrule
    \rowcolor{gray!15}\multicolumn{4}{l}{\textbf{Full Fine-tuned SLMs with full dataset}} \\
    \multicolumn{1}{l}{GenDA} & 50.01  & 60.88  & \cellcolor{gray!15}55.45  \\
    \multicolumn{1}{l}{LEGO-ABSA} &  46.10  & 57.60  & \cellcolor{gray!15}51.85  \\
    \multicolumn{1}{l}{MVP} &  \textbf{52.21}  & \textbf{58.94}  & \cellcolor{gray!15}\textbf{55.58}  \\
    \multicolumn{1}{l}{T5-Instruct} &  48.54 &	54.22 &	\cellcolor{gray!15}51.38  \\
    \rowcolor{gray!15}\multicolumn{4}{l}{\textbf{Efficiently Fine-tuned LLMs with full dataset}} \\
    ChatGLM3-6B &  47.90  & 53.16  & \cellcolor{gray!15}50.53  \\
    QWen1.5-7B   & 53.45  & 58.23  & \cellcolor{gray!15}55.84  \\
    Mistral-7B-v0.2 & 52.91  & 57.94  & \cellcolor{gray!15}55.43  \\
    LLaMA3-8B  & \textbf{57.07}  & \textbf{58.87}  & \cellcolor{gray!15}\textbf{57.97}  \\
    \midrule[1pt]
    \rowcolor{gray!15}\multicolumn{4}{l}{\textbf{API-based LLMs without fine-tuning }} \\
    \rowcolor{gray!15}\multicolumn{4}{l}{\textbf{(zero-/three-shot)}} \\
    
    LLaMA3-70B & 15.31  & 16.26  & \cellcolor{gray!15}15.79  \\
    \textit{\ \ +Random} & 28.71  & 32.93  & \cellcolor{gray!15}30.82  \\
    \textit{\ \ +BM25} &  36.94  & 43.10  & \cellcolor{gray!15}40.02  \\
    \textit{\ \ +SimCSE} &  35.71  & 42.59  & \cellcolor{gray!15}39.15  \\
    \cmidrule{2-4}
    ChatGPT & 15.23  & 16.49  & \cellcolor{gray!15}15.86  \\
    \textit{\ \ +Random} &  25.40  & 26.06  & \cellcolor{gray!15}25.73  \\
    \textit{\ \ +BM25} &  31.34  & 36.73  & \cellcolor{gray!15}34.04  \\
    \textit{\ \ +SimCSE} &  32.37  & 37.35  & \cellcolor{gray!15}34.86  \\
    \cmidrule{2-4}
    GPT4+\textit{BM25} & \textbf{37.84}  & \textbf{43.73}  & \cellcolor{gray!15}\textbf{40.79}  \\
    \bottomrule[1.2pt]
    \end{tabular}%
    \caption{Main results of ASQP subtask.}
  \label{exp:main3}%
\end{table}%

\subsubsection{Implement Details}

We use \verb|rank_bm25| library~\footnote{\url{https://github.com/dorianbrown/rank_bm25}} to execute BM25 retrieval, and use \verb|sup-simcse-roberta-large| to execute SimCSE retrieval. We use three-shot as default ICL setting, and utilize \verb|LLaMA-Factory|~\cite{zheng2024llamafactory}, a easy and efficient LLM fine-tuning library~\footnote{\url{https://github.com/hiyouga/LLaMA-Factory}}, to fine-tune all open source LLMs with LoRA. All weights of LLMs come from HuggingFace~\footnote{\url{https://huggingface.co/models}}. All experiments run on an Ubuntu Server with 8 Nvidia RTX 3090 GPUs. 


\subsubsection{Baselines}

\textbf{(1) Full Fine-tuned SLMs}: A variety of methods have been proposed for different ABSA subtasks~\cite{zhang2023survey}. Here, we list state-of-the-art methods that target one or more of these subtasks: \textbf{Dual-MRC}~\cite{Mao_Shen_Yu_Cai_2021}, \textbf{DCRAN}~\cite{oh-etal-2021-deep}, \textbf{BARTABSA}~\cite{yan-etal-2021-unified}, \textbf{LEGO-ABSA}~\cite{gao-etal-2022-lego}, \textbf{TAGS}~\cite{xianlong-etal-2023-tagging}, \textbf{GenDA}~\cite{wang-etal-2023-generative}, and \textbf{MVP}~\cite{gou-etal-2023-mvp}. Besides, we conduct instruction-based multi-task fine-tuning on the T5-base model using unified task formulation mentioned in the Methodology section, namely \textbf{T5-Instruct}. \textbf{(2) Efficient Fine-tuned LLMs}: We use LoRA to efficiently fine-tune the following open-source LLMs: \textbf{LLaMA3-8B}~\footnote{All open source LLMs mentioned in this paper are Instruction version. We omit ``Instruction'' for convenience.}~\cite{llama3}, \textbf{ChatGLM3-6B}~\cite{du-etal-2022-glm, zeng2023glm130b}, \textbf{QWen1.5-7B}~\cite{bai2023qwen}, and \textbf{Mistral-7B-v0.2}~\cite{jiang2023mistral}. \textbf{(3) API-based LLMs}: We evaluate the zero-shot and few-shot abilities of the following LLMs by calling the APIs: \textbf{LLaMA3-70B}, \textbf{ChatGPT}~\cite{chatgpt} (\verb|GPT-3.5-Turbo-0125|), and \textbf{GPT4}~\cite{openai2024gpt4} (\verb|GPT-4-Turbo|). 



\subsection{Main Results}
\label{mainresults}

Due to space constraints, we divide ABSA subtasks into three groups, as shown in Tables~\ref{exp:main1}, ~\ref{exp:main2} and ~\ref{exp:main3}. These tables show the F1 scores of SLMs and LLMs under the fine-tuning-dependent paradigm, as well as the F1 scores of LLMs under the fine-tuning-free paradigm. The results for \textbf{SLMs without fine-tuning} are not displayed due to their ineffectiveness in all subtasks, with F1 scores close to 0.

\begin{figure*}[t]
		\centering
\includegraphics[width=1.0\linewidth]{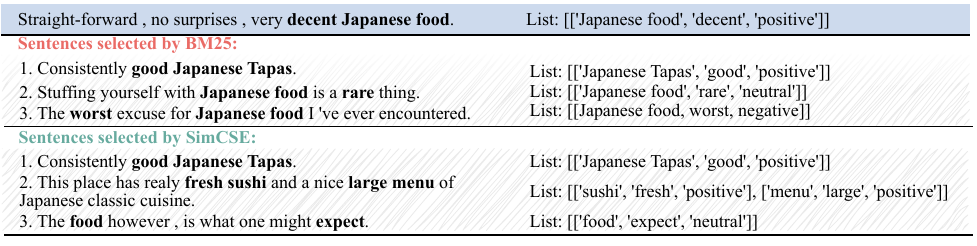}
		\caption{An example of BM25-based and SimCSE-based selection strategies for ASTE subtask.
		}
		\label{fur:fig1}
\end{figure*}

\subsubsection{Full Fine-tuned SLMs vs. Efficient Fine-tuned LLMs}
\label{ellms}
We efficiently fine-tune four open source LLMs with LoRA by instruction-based multi-task learning. Thanks to the powerful natural language understanding and generation abilities, LLMs achieve state-of-the-art performance with only fine-tuning minimal parameters (e.g., LLaMA3-8B with 3.4M vs. T5-Instruct with 220M). The sentiment analysis and structured extraction abilities stimulated by LoRA enable LLMs outperform SLMs 2.87, 4.67, and 2.39 points across three groups of subtasks, respectively. This indicates that LLMs are a better choice than SLMs when are fine-tuned with sufficient data.

\begin{figure}[t]
	\centering
    \includegraphics[width=1.0\columnwidth]{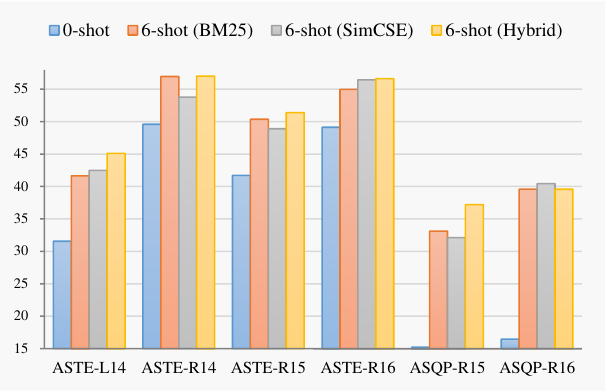}
	\caption{The F1 scores of ChatGPT on ASTE and ASQP subtasks. The ``Hybrid'' denotes combining three demonstrations selected by BM25 and three demonstrations selected by SimCSE, arranged in random order.
		}
		\label{result:icl}
\end{figure}

\subsubsection{Zero-shot and Few-shot Abilities of LLMs}
\label{effectoficl}

SLMs are ineffective in the fine-tuning-free paradigm due to their weak natural language understanding and instruction-following abilities. In contrast, LLMs exhibit surprisingly strong zero-shot performance, even rivaling fine-tuned SLMs in the ALSC subtask. This demonstrates that LLMs can comprehend the nature of a subtask and achieve it based on instructional descriptions. When using in-context learning (ICL), the performance of LLMs further improves, indicating they can learn the annotation rules of a subtask from demonstrations. Besides, GPT4~\footnote{We only evaluate the performance in BM25 selection strategy due to the expensive price of GPT4.}, as the currently strongest LLM, remains unbeatable. Overall, the performance of LLMs with ICL demonstrates that they can be effectively utilized in low-resource fine-tuning-free scenarios.

\begin{figure}[t]
	\centering
    \includegraphics[width=1.0\linewidth]{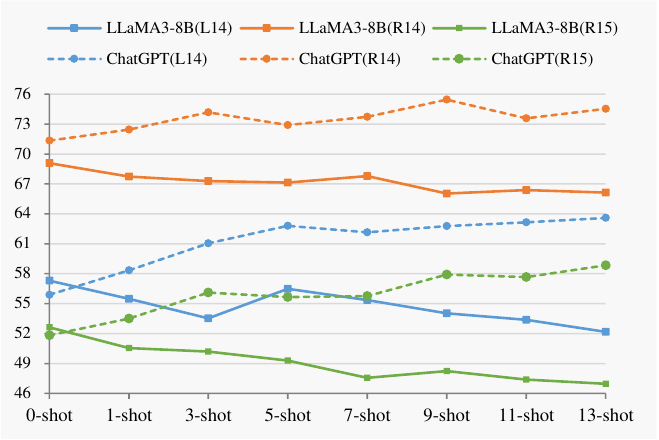}
	\caption{The F1 scores of LLaMA3-8B+\textit{Random} and ChatGPT+\textit{Random} on the OE subtask.}
	\label{fur:fig3}
\end{figure}


\subsubsection{Demonstration Selection Strategies}

The improvement brought by random demonstrations is limited. Even these demonstrations may cause negative effect, as shown by the underlined entries in Tables 2 and 3. This is because there is no correlation between the selected demonstrations and the tested samples. In contrast, the selection strategies based on BM25 and SimCSE can achieve comparable performance enhancements, and both outperform the random strategy. This indicates that the importance of keyword and semantic information is roughly equivalent. We show an example of two selection strategies for the ASTE subtask in Figure~\ref{fur:fig1}. As mentioned earlier in the Methodology section, BM25 can retrieve sentences with keywords similar to those in the target sentence, which is helpful for extracting aspect terms. SimCSE can retrieve sentences that have semantic relevance to the target sentence. These sentences express slightly positive comments about food, which promotes semantic understanding and sentiment recognition of the target sentence. Furthermore, we study whether both types of keyword and semantic information can be simultaneously effective. The results are shown in Figure~\ref{result:icl}, the Hybrid strategy achieves the better performance than BM25 and SimCSE, indicating the effectiveness and complementarity of keyword and semantic information.

\subsubsection{Failed ICL}

As mentioned earlier, in rare cases, ICL can deteriorate the performance of LLMs, as shown by the underlined entries in Tables~\ref{exp:main1} and~\ref{exp:main2}. This phenomenon is primarily evident with the LLaMA3-70B using random demonstrations in the relatively simple subtasks, which indicates LLaMA3-70B are more ``picky'' with demonstrations compared to ChatGPT, despite it shows stronger basic zero-shot abilities than ChatGPT. This also indicates that additional demonstrations increase the prompt length and degrade the performance in simple subtasks. 

We further study this phenomenon by changing the number of demonstrations on the OE subtask using LLaMA3-8B+\textit{Random} and ChatGPT+\textit{Random}. As shown in Figure~\ref{fur:fig3}, with the increase of the number of demonstrations, the F1 scores of ChatGPT (dashed line) first rise and then fluctuate, which indicates that an appropriate increase in the number of demonstrations is effective. In contrast, for the LLaMA3-8B model (solid line), increasing the number of demonstrations not only fails to improve performance but also exacerbates its decline. This phenomenon indicates that more demonstrations cannot reverse the trend of performance decline. In other words, the degraded performance of LLaMA3-8B and LLaMA3-70B is inherent to the model rather than due to insufficient number of demonstrations.

\subsection{Further Study}
\label{furtherstudy}

\begin{figure*}[t]
	\centering
\includegraphics[width=0.9\linewidth]{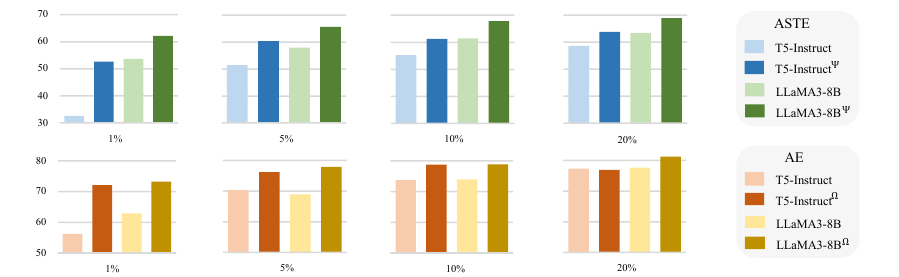}
	\caption{The above and below figures show the average F1 scores on the ASTE and AE subtasks in low-resource fine-tuning settings (1\%, 5\%, 10\% and 20\% of training datas), respectively. $\Psi$ denotes that before training with minimal data for the ASTE subtask, the model will first be trained on full data of \{AE, OE, ALSC, AOE\} subtasks. Similarly, $\Omega$ denote that before training with minimal data for the AE subtask, the model will first be trained on full data of \{AESC, AOPE, ASTE, ASQP\} subtasks.}
    \label{fur:final}
\end{figure*}

\setlength{\tabcolsep}{6pt}
\begin{table}[htbp]
  \centering
    \begin{tabular}{lcccc}
    \toprule
    Fine-Tuning      & ALSC  & AOPE  & ASTE  & ASQP \\
    \midrule
    LLaMA3-8B & \textbf{86.16}  & \textbf{79.37}  & 73.77  & 57.97  \\
    \ \ +\textit{Random} & 85.88  & 78.70  & \textbf{75.06}  & \textbf{58.76}  \\
    \ \ +\textit{BM25}  & 86.01  & 78.64  & 73.39  & 56.36  \\
    \ \ +\textit{SimCSE} & 85.93  & 78.24  & 73.17  & 58.36  \\
    \bottomrule
    \end{tabular}%
    \caption{The average F1 scores in the four subtasks. ``+'' denotes in-context fine-tuning setting.}
  \label{fur:tab1}%
\end{table}%

\subsubsection{In-context Fine-tuning}

Researchers propose in-context fine-tuning method by constructing in-context training data to improve ICL ability of LLMs~\cite{dong2023survey}. To this end, we select three demonstrations for each training sample from the training set itself. The results on the four subtasks are shown in Table~\ref{fur:tab1}. For the two relatively simple ALSC and AOPE subtasks, random demonstrations cannot bring improvement for the training of LLMs. This is because LLMs can understand simple subtasks only by instruction descriptions and backpropagation of gradients. In contrast, for complex ASTE and ASQP subtasks, random demonstrations can improve performance by aiding LLMs in learning annoatation rules. Besides, BM25 and SimCSE that perform well in fine-tuning-free paradigm become ineffective. We assume that these two strategies cause LLMs to take shortcuts in fine-tuning, leading them to rely excessively on keyword or semantic information rather than genuinely understanding the task itself.


\begin{figure}[t]
	\centering
    \includegraphics[width=0.9\columnwidth]{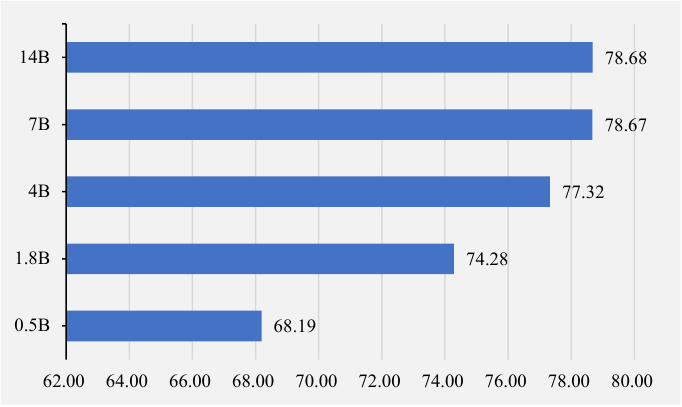}
	\caption{The average F1 scores of QWen1.5 on all subtasks.
		}
		\label{fur:fig2}
\end{figure}

\subsubsection{Cross-task Transfer in Low-resource Fine-tuning}


In this section, we explore the performance of cross-task transfer of LLMs and SLMs in low-resource fine-tuning settings, as shown in Figure~\ref{fur:final}. For the ASTE subtask, warming up on four simple subtasks can improve the performance of models, especially when using 1\% of training data. Similarly, warming up on complex subtasks improve the performance of the AE subtask. This illustrates that subtasks can be mutually enhanced. When faced with a new subtask where data is scarce, warming up on existing subtasks makes sense. Besides, the performance of LLMs is better than that of SLMs, whether warming up is used or not, demonstrating the advantages of LLMs in low-resource training settings.

\subsubsection{Effect of Parameter Size} 

We investigate the impact of parameter size of fine-tuned LLMs using five versions of the QWen1.5 model.
As shown in Figure~\ref{fur:fig2}, the parameter size demonstrates a diminishing returns effect on ABSA subtasks. This suggests that for ABSA subtasks, LLMs with around 7B parameters are the optimal choice. Models that are too small lack instruction-following and extraction abilities, while models that are too large bring limited improvements and significantly increase training and inference costs.





\section{Conclusion}

This paper evaluates the performance of 6 LLMs on 13 datasets of 8 ABSA subtasks, and unify ``multiple LLMs for multiple subtasks in multiple paradigms.'' Comprehensive experiments demonstrate that the performance of LLMs is better than that of SLMs, regardless of fine-tuning. For fine-tuning-dependent paradigm, fine-tuned LLMs with LoRA obtain cost-effective and state-of-the-art performance, surpassing SLMs on all subtasks. For fine-tuning-free paradigm, although it is not yet comparable to fine-tuned models, LLMs with ICL shed light on the possibility of achieving ABSA in low-resource scenarios. Besides, the effectiveness of ICL varies across different ABSA subtasks, LLM, and demonstration selection strategies. In rare cases, ICL may fail. In summary, our comprehensive study demonstrates that the overall success of LLMs over SLMs. 


\bibliography{aaai25}

\begin{thebibliography}{53}
\providecommand{\natexlab}[1]{#1}

\bibitem[{Amin et~al.(2023)Amin, Mao, Cambria, and Schuller}]{amin2023wide}
Amin, M.~M.; Mao, R.; Cambria, E.; and Schuller, B.~W. 2023.
\newblock A Wide Evaluation of ChatGPT on Affective Computing Tasks.
\newblock arXiv:2308.13911.

\bibitem[{Bai et~al.(2023)Bai, Bai, Chu, Cui, Dang, Deng, Fan, Ge, Han, Huang, Hui, Ji et~al.}]{bai2023qwen}
Bai, J.; Bai, S.; Chu, Y.; Cui, Z.; Dang, K.; Deng, X.; Fan, Y.; Ge, W.; Han, Y.; Huang, F.; Hui, B.; Ji, L.; et~al. 2023.
\newblock Qwen Technical Report.
\newblock arXiv:2309.16609.

\bibitem[{Brown et~al.(2020)Brown, Mann, Ryder, Subbiah, Kaplan, Dhariwal, Neelakantan, Shyam, Sastry, Askell, Agarwal, Herbert-Voss, Krueger, Henighan, Child, Ramesh, Ziegler, Wu, Winter, Hesse, Chen, Sigler, Litwin, Gray, Chess, Clark, Berner, McCandlish, Radford, Sutskever, and Amodei}]{brown2020language}
Brown, T.~B.; Mann, B.; Ryder, N.; Subbiah, M.; Kaplan, J.; Dhariwal, P.; Neelakantan, A.; Shyam, P.; Sastry, G.; Askell, A.; Agarwal, S.; Herbert-Voss, A.; Krueger, G.; Henighan, T.; Child, R.; Ramesh, A.; Ziegler, D.~M.; Wu, J.; Winter, C.; Hesse, C.; Chen, M.; Sigler, E.; Litwin, M.; Gray, S.; Chess, B.; Clark, J.; Berner, C.; McCandlish, S.; Radford, A.; Sutskever, I.; and Amodei, D. 2020.
\newblock Language Models are Few-Shot Learners.
\newblock arXiv:2005.14165.

\bibitem[{Cai, Xia, and Yu(2021)}]{cai-etal-2021-aspect}
Cai, H.; Xia, R.; and Yu, J. 2021.
\newblock Aspect-Category-Opinion-Sentiment Quadruple Extraction with Implicit Aspects and Opinions.
\newblock In Zong, C.; Xia, F.; Li, W.; and Navigli, R., eds., \emph{Proceedings of the 59th Annual Meeting of the Association for Computational Linguistics and the 11th International Joint Conference on Natural Language Processing (Volume 1: Long Papers)}, 340--350. Online: Association for Computational Linguistics.

\bibitem[{Dong et~al.(2023)Dong, Li, Dai, Zheng, Wu, Chang, Sun, Xu, Li, and Sui}]{dong2023survey}
Dong, Q.; Li, L.; Dai, D.; Zheng, C.; Wu, Z.; Chang, B.; Sun, X.; Xu, J.; Li, L.; and Sui, Z. 2023.
\newblock A Survey on In-context Learning.
\newblock arXiv:2301.00234.

\bibitem[{Du et~al.(2022)Du, Qian, Liu, Ding, Qiu, Yang, and Tang}]{du-etal-2022-glm}
Du, Z.; Qian, Y.; Liu, X.; Ding, M.; Qiu, J.; Yang, Z.; and Tang, J. 2022.
\newblock {GLM}: General Language Model Pretraining with Autoregressive Blank Infilling.
\newblock In Muresan, S.; Nakov, P.; and Villavicencio, A., eds., \emph{Proceedings of the 60th Annual Meeting of the Association for Computational Linguistics (Volume 1: Long Papers)}, 320--335. Dublin, Ireland: Association for Computational Linguistics.

\bibitem[{Fan et~al.(2019)Fan, Wu, Dai, Huang, and Chen}]{fan-etal-2019-target}
Fan, Z.; Wu, Z.; Dai, X.-Y.; Huang, S.; and Chen, J. 2019.
\newblock Target-oriented Opinion Words Extraction with Target-fused Neural Sequence Labeling.
\newblock In Burstein, J.; Doran, C.; and Solorio, T., eds., \emph{Proceedings of the 2019 Conference of the North {A}merican Chapter of the Association for Computational Linguistics: Human Language Technologies, Volume 1 (Long and Short Papers)}, 2509--2518. Minneapolis, Minnesota: Association for Computational Linguistics.

\bibitem[{Gao et~al.(2022)Gao, Fang, Liu, Liu, Liu, Liu, Bao, and Yan}]{gao-etal-2022-lego}
Gao, T.; Fang, J.; Liu, H.; Liu, Z.; Liu, C.; Liu, P.; Bao, Y.; and Yan, W. 2022.
\newblock {LEGO}-{ABSA}: A Prompt-based Task Assemblable Unified Generative Framework for Multi-task Aspect-based Sentiment Analysis.
\newblock In Calzolari, N.; Huang, C.-R.; Kim, H.; Pustejovsky, J.; Wanner, L.; Choi, K.-S.; Ryu, P.-M.; Chen, H.-H.; Donatelli, L.; Ji, H.; Kurohashi, S.; Paggio, P.; Xue, N.; Kim, S.; Hahm, Y.; He, Z.; Lee, T.~K.; Santus, E.; Bond, F.; and Na, S.-H., eds., \emph{Proceedings of the 29th International Conference on Computational Linguistics}, 7002--7012. Gyeongju, Republic of Korea: International Committee on Computational Linguistics.

\bibitem[{Gao, Yao, and Chen(2021)}]{gao-etal-2021-simcse}
Gao, T.; Yao, X.; and Chen, D. 2021.
\newblock {S}im{CSE}: Simple Contrastive Learning of Sentence Embeddings.
\newblock In Moens, M.-F.; Huang, X.; Specia, L.; and Yih, S. W.-t., eds., \emph{Proceedings of the 2021 Conference on Empirical Methods in Natural Language Processing}, 6894--6910. Online and Punta Cana, Dominican Republic: Association for Computational Linguistics.

\bibitem[{Gao et~al.(2024)Gao, Xiong, Gao, Jia, Pan, Bi, Dai, Sun, Wang, and Wang}]{gao2024retrievalaugmented}
Gao, Y.; Xiong, Y.; Gao, X.; Jia, K.; Pan, J.; Bi, Y.; Dai, Y.; Sun, J.; Wang, M.; and Wang, H. 2024.
\newblock Retrieval-Augmented Generation for Large Language Models: A Survey.
\newblock arXiv:2312.10997.

\bibitem[{Gou, Guo, and Yang(2023)}]{gou-etal-2023-mvp}
Gou, Z.; Guo, Q.; and Yang, Y. 2023.
\newblock {M}v{P}: Multi-view Prompting Improves Aspect Sentiment Tuple Prediction.
\newblock In Rogers, A.; Boyd-Graber, J.; and Okazaki, N., eds., \emph{Proceedings of the 61st Annual Meeting of the Association for Computational Linguistics (Volume 1: Long Papers)}, 4380--4397. Toronto, Canada: Association for Computational Linguistics.

\bibitem[{Guo et~al.(2024)Guo, Chen, Wang, Chang, Pei, Chawla, Wiest, and Zhang}]{guo2024large}
Guo, T.; Chen, X.; Wang, Y.; Chang, R.; Pei, S.; Chawla, N.~V.; Wiest, O.; and Zhang, X. 2024.
\newblock Large Language Model based Multi-Agents: A Survey of Progress and Challenges.
\newblock arXiv:2402.01680.

\bibitem[{Han et~al.(2024)Han, Gao, Liu, Zhang, and Zhang}]{han2024peftsurvey}
Han, Z.; Gao, C.; Liu, J.; Zhang, J.; and Zhang, S.~Q. 2024.
\newblock Parameter-Efficient Fine-Tuning for Large Models: A Comprehensive Survey.
\newblock arXiv:2403.14608.

\bibitem[{Hu et~al.(2022)Hu, yelong shen, Wallis, Allen-Zhu, Li, Wang, Wang, and Chen}]{hu2022lora}
Hu, E.~J.; yelong shen; Wallis, P.; Allen-Zhu, Z.; Li, Y.; Wang, S.; Wang, L.; and Chen, W. 2022.
\newblock Lo{RA}: Low-Rank Adaptation of Large Language Models.
\newblock In \emph{International Conference on Learning Representations}.

\bibitem[{Jiang et~al.(2023)Jiang, Sablayrolles, Mensch, Bamford, Chaplot, de~las Casas, Bressand, Lengyel, Lample, Saulnier, Lavaud, Lachaux, Stock, Scao, Lavril, Wang, Lacroix, and Sayed}]{jiang2023mistral}
Jiang, A.~Q.; Sablayrolles, A.; Mensch, A.; Bamford, C.; Chaplot, D.~S.; de~las Casas, D.; Bressand, F.; Lengyel, G.; Lample, G.; Saulnier, L.; Lavaud, L.~R.; Lachaux, M.-A.; Stock, P.; Scao, T.~L.; Lavril, T.; Wang, T.; Lacroix, T.; and Sayed, W.~E. 2023.
\newblock Mistral 7B.
\newblock arXiv:2310.06825.

\bibitem[{Lewis et~al.(2020)Lewis, Liu, Goyal, Ghazvininejad, Mohamed, Levy, Stoyanov, and Zettlemoyer}]{lewis-etal-2020-bart}
Lewis, M.; Liu, Y.; Goyal, N.; Ghazvininejad, M.; Mohamed, A.; Levy, O.; Stoyanov, V.; and Zettlemoyer, L. 2020.
\newblock {BART}: Denoising Sequence-to-Sequence Pre-training for Natural Language Generation, Translation, and Comprehension.
\newblock In Jurafsky, D.; Chai, J.; Schluter, N.; and Tetreault, J., eds., \emph{Proceedings of the 58th Annual Meeting of the Association for Computational Linguistics}, 7871--7880. Online: Association for Computational Linguistics.

\bibitem[{Li, Wang, and Ke(2023)}]{li-etal-2023-revisiting-large}
Li, G.; Wang, P.; and Ke, W. 2023.
\newblock Revisiting Large Language Models as Zero-shot Relation Extractors.
\newblock In Bouamor, H.; Pino, J.; and Bali, K., eds., \emph{Findings of the Association for Computational Linguistics: EMNLP 2023}, 6877--6892. Singapore: Association for Computational Linguistics.

\bibitem[{Liu(2012)}]{liu2012sentiment}
Liu, B. 2012.
\newblock Sentiment analysis and opinion mining.
\newblock \emph{Synthesis lectures on human language technologies}, 5(1): 1--167.

\bibitem[{Mao et~al.(2022)Mao, Shen, Yang, Zhu, and Cai}]{mao-etal-2022-seq2path}
Mao, Y.; Shen, Y.; Yang, J.; Zhu, X.; and Cai, L. 2022.
\newblock {S}eq2{P}ath: Generating Sentiment Tuples as Paths of a Tree.
\newblock In Muresan, S.; Nakov, P.; and Villavicencio, A., eds., \emph{Findings of the Association for Computational Linguistics: ACL 2022}, 2215--2225. Dublin, Ireland: Association for Computational Linguistics.

\bibitem[{Mao et~al.(2021)Mao, Shen, Yu, and Cai}]{Mao_Shen_Yu_Cai_2021}
Mao, Y.; Shen, Y.; Yu, C.; and Cai, L. 2021.
\newblock A Joint Training Dual-MRC Framework for Aspect Based Sentiment Analysis.
\newblock \emph{Proceedings of the AAAI Conference on Artificial Intelligence}, 35(15): 13543--13551.

\bibitem[{MetaAI(2024)}]{llama3}
MetaAI. 2024.
\newblock Introducing Meta Llama 3: The most capable openly available LLM to date.

\bibitem[{Oh et~al.(2021)Oh, Lee, Whang, Park, Gaeun, Kim, and Kim}]{oh-etal-2021-deep}
Oh, S.; Lee, D.; Whang, T.; Park, I.; Gaeun, S.; Kim, E.; and Kim, H. 2021.
\newblock Deep Context- and Relation-Aware Learning for Aspect-based Sentiment Analysis.
\newblock In Zong, C.; Xia, F.; Li, W.; and Navigli, R., eds., \emph{Proceedings of the 59th Annual Meeting of the Association for Computational Linguistics and the 11th International Joint Conference on Natural Language Processing (Volume 2: Short Papers)}, 495--503. Online: Association for Computational Linguistics.

\bibitem[{OpenAI(2022)}]{chatgpt}
OpenAI. 2022.
\newblock Introducing ChatGPT.

\bibitem[{OpenAI(2023)}]{openai2024gpt4}
OpenAI. 2023.
\newblock GPT-4 Technical Report.
\newblock arXiv:2303.08774.

\bibitem[{Peng et~al.(2020)Peng, Xu, Bing, Huang, Lu, and Si}]{Peng_Xu_Bing_Huang_Lu_Si_2020}
Peng, H.; Xu, L.; Bing, L.; Huang, F.; Lu, W.; and Si, L. 2020.
\newblock Knowing What, How and Why: A Near Complete Solution for Aspect-Based Sentiment Analysis.
\newblock \emph{Proceedings of the AAAI Conference on Artificial Intelligence}, 34(05): 8600--8607.

\bibitem[{Pontiki et~al.(2016)Pontiki, Galanis, Papageorgiou, Androutsopoulos, Manandhar, AL-Smadi, Al-Ayyoub, Zhao, Qin, De~Clercq, Hoste, Apidianaki, Tannier, Loukachevitch, Kotelnikov, Bel, Jim{\'e}nez-Zafra, and Eryi{\u{g}}it}]{pontiki-etal-2016-semeval}
Pontiki, M.; Galanis, D.; Papageorgiou, H.; Androutsopoulos, I.; Manandhar, S.; AL-Smadi, M.; Al-Ayyoub, M.; Zhao, Y.; Qin, B.; De~Clercq, O.; Hoste, V.; Apidianaki, M.; Tannier, X.; Loukachevitch, N.; Kotelnikov, E.; Bel, N.; Jim{\'e}nez-Zafra, S.~M.; and Eryi{\u{g}}it, G. 2016.
\newblock {S}em{E}val-2016 Task 5: Aspect Based Sentiment Analysis.
\newblock In Bethard, S.; Carpuat, M.; Cer, D.; Jurgens, D.; Nakov, P.; and Zesch, T., eds., \emph{Proceedings of the 10th International Workshop on Semantic Evaluation ({S}em{E}val-2016)}, 19--30. San Diego, California: Association for Computational Linguistics.

\bibitem[{Pontiki et~al.(2015)Pontiki, Galanis, Papageorgiou, Manandhar, and Androutsopoulos}]{pontiki-etal-2015-semeval}
Pontiki, M.; Galanis, D.; Papageorgiou, H.; Manandhar, S.; and Androutsopoulos, I. 2015.
\newblock {S}em{E}val-2015 Task 12: Aspect Based Sentiment Analysis.
\newblock In Nakov, P.; Zesch, T.; Cer, D.; and Jurgens, D., eds., \emph{Proceedings of the 9th International Workshop on Semantic Evaluation ({S}em{E}val 2015)}, 486--495. Denver, Colorado: Association for Computational Linguistics.

\bibitem[{Pontiki et~al.(2014)Pontiki, Galanis, Pavlopoulos, Papageorgiou, Androutsopoulos, and Manandhar}]{pontiki-etal-2014-semeval}
Pontiki, M.; Galanis, D.; Pavlopoulos, J.; Papageorgiou, H.; Androutsopoulos, I.; and Manandhar, S. 2014.
\newblock {S}em{E}val-2014 Task 4: Aspect Based Sentiment Analysis.
\newblock In Nakov, P.; and Zesch, T., eds., \emph{Proceedings of the 8th International Workshop on Semantic Evaluation ({S}em{E}val 2014)}, 27--35. Dublin, Ireland: Association for Computational Linguistics.

\bibitem[{Raffel et~al.(2020)Raffel, Shazeer, Roberts, Lee, Narang, Matena, Zhou, Li, and Liu}]{t5}
Raffel, C.; Shazeer, N.; Roberts, A.; Lee, K.; Narang, S.; Matena, M.; Zhou, Y.; Li, W.; and Liu, P.~J. 2020.
\newblock Exploring the Limits of Transfer Learning with a Unified Text-to-Text Transformer.
\newblock \emph{Journal of Machine Learning Research}, 21(140): 1--67.

\bibitem[{Robertson and Zaragoza(2009)}]{2009bm25}
Robertson, S.; and Zaragoza, H. 2009.
\newblock The Probabilistic Relevance Framework: BM25 and Beyond.
\newblock \emph{Found. Trends Inf. Retr.}, 3(4): 333–389.

\bibitem[{Simmering and Huoviala(2023)}]{simmering2023large}
Simmering, P.~F.; and Huoviala, P. 2023.
\newblock Large language models for aspect-based sentiment analysis.
\newblock arXiv:2310.18025.

\bibitem[{Vinyals, Fortunato, and Jaitly(2015)}]{NIPS2015_29921001}
Vinyals, O.; Fortunato, M.; and Jaitly, N. 2015.
\newblock Pointer Networks.
\newblock In Cortes, C.; Lawrence, N.; Lee, D.; Sugiyama, M.; and Garnett, R., eds., \emph{Advances in Neural Information Processing Systems}, volume~28. Curran Associates, Inc.

\bibitem[{Wadhwa, Amir, and Wallace(2023)}]{wadhwa-etal-2023-revisiting}
Wadhwa, S.; Amir, S.; and Wallace, B. 2023.
\newblock Revisiting Relation Extraction in the era of Large Language Models.
\newblock In Rogers, A.; Boyd-Graber, J.; and Okazaki, N., eds., \emph{Proceedings of the 61st Annual Meeting of the Association for Computational Linguistics (Volume 1: Long Papers)}, 15566--15589. Toronto, Canada: Association for Computational Linguistics.

\bibitem[{Wan et~al.(2020)Wan, Yang, Du, Liu, Qi, and Pan}]{Wan_Yang_Du_Liu_Qi_Pan_2020}
Wan, H.; Yang, Y.; Du, J.; Liu, Y.; Qi, K.; and Pan, J.~Z. 2020.
\newblock Target-Aspect-Sentiment Joint Detection for Aspect-Based Sentiment Analysis.
\newblock \emph{Proceedings of the AAAI Conference on Artificial Intelligence}, 34(05): 9122--9129.

\bibitem[{Wang et~al.(2023{\natexlab{a}})Wang, Jiang, Ma, Liu, and Okazaki}]{wang-etal-2023-generative}
Wang, A.; Jiang, J.; Ma, Y.; Liu, A.; and Okazaki, N. 2023{\natexlab{a}}.
\newblock Generative Data Augmentation for Aspect Sentiment Quad Prediction.
\newblock In Palmer, A.; and Camacho-collados, J., eds., \emph{Proceedings of the 12th Joint Conference on Lexical and Computational Semantics (*SEM 2023)}, 128--140. Toronto, Canada: Association for Computational Linguistics.

\bibitem[{Wang et~al.(2023{\natexlab{b}})Wang, Sun, Li, Ouyang, Wu, Zhang, Li, and Wang}]{wang2023gptner}
Wang, S.; Sun, X.; Li, X.; Ouyang, R.; Wu, F.; Zhang, T.; Li, J.; and Wang, G. 2023{\natexlab{b}}.
\newblock GPT-NER: Named Entity Recognition via Large Language Models.
\newblock arXiv:2304.10428.

\bibitem[{Wang et~al.(2017)Wang, Pan, Dahlmeier, and Xiao}]{Wang_Pan_Dahlmeier_Xiao_2017}
Wang, W.; Pan, S.~J.; Dahlmeier, D.; and Xiao, X. 2017.
\newblock Coupled Multi-Layer Attentions for Co-Extraction of Aspect and Opinion Terms.
\newblock \emph{Proceedings of the AAAI Conference on Artificial Intelligence}, 31(1).

\bibitem[{Wang, Xia, and Yu(2024)}]{wang2024unifiedabsa}
Wang, Z.; Xia, R.; and Yu, J. 2024.
\newblock Unified ABSA via Annotation-Decoupled Multi-task Instruction Tuning.
\newblock \emph{IEEE Transactions on Knowledge and Data Engineering}, 1--13.

\bibitem[{Wang et~al.(2024)Wang, Xie, Feng, Ding, Yang, and Xia}]{wang2024chatgpt}
Wang, Z.; Xie, Q.; Feng, Y.; Ding, Z.; Yang, Z.; and Xia, R. 2024.
\newblock Is ChatGPT a Good Sentiment Analyzer? A Preliminary Study.
\newblock arXiv:2304.04339.

\bibitem[{Xianlong, Yang, and Wang(2023)}]{xianlong-etal-2023-tagging}
Xianlong, L.; Yang, M.; and Wang, Y. 2023.
\newblock Tagging-Assisted Generation Model with Encoder and Decoder Supervision for Aspect Sentiment Triplet Extraction.
\newblock In Bouamor, H.; Pino, J.; and Bali, K., eds., \emph{Proceedings of the 2023 Conference on Empirical Methods in Natural Language Processing}, 2078--2093. Singapore: Association for Computational Linguistics.

\bibitem[{Xu et~al.(2024)Xu, Chen, Peng, Zhang, Xu, Zhao, Wu, Zheng, Wang, and Chen}]{xu2024large}
Xu, D.; Chen, W.; Peng, W.; Zhang, C.; Xu, T.; Zhao, X.; Wu, X.; Zheng, Y.; Wang, Y.; and Chen, E. 2024.
\newblock Large Language Models for Generative Information Extraction: A Survey.
\newblock arXiv:2312.17617.

\bibitem[{Xu et~al.(2023)Xu, Zhang, Xiao, and Xiong}]{xu2023limits}
Xu, X.; Zhang, J.-D.; Xiao, R.; and Xiong, L. 2023.
\newblock The Limits of ChatGPT in Extracting Aspect-Category-Opinion-Sentiment Quadruples: A Comparative Analysis.
\newblock arXiv:2310.06502.

\bibitem[{Yan et~al.(2021)Yan, Dai, Ji, Qiu, and Zhang}]{yan-etal-2021-unified}
Yan, H.; Dai, J.; Ji, T.; Qiu, X.; and Zhang, Z. 2021.
\newblock A Unified Generative Framework for Aspect-based Sentiment Analysis.
\newblock In Zong, C.; Xia, F.; Li, W.; and Navigli, R., eds., \emph{Proceedings of the 59th Annual Meeting of the Association for Computational Linguistics and the 11th International Joint Conference on Natural Language Processing (Volume 1: Long Papers)}, 2416--2429. Online: Association for Computational Linguistics.

\bibitem[{Zeng et~al.(2023)Zeng, Liu, Du, Wang, Lai, Ding, Yang, Xu, Zheng, Xia, Tam, Ma, Xue, Zhai, Chen, Zhang, Dong, and Tang}]{zeng2023glm130b}
Zeng, A.; Liu, X.; Du, Z.; Wang, Z.; Lai, H.; Ding, M.; Yang, Z.; Xu, Y.; Zheng, W.; Xia, X.; Tam, W.~L.; Ma, Z.; Xue, Y.; Zhai, J.; Chen, W.; Zhang, P.; Dong, Y.; and Tang, J. 2023.
\newblock GLM-130B: An Open Bilingual Pre-trained Model.
\newblock arXiv:2210.02414.

\bibitem[{Zhang et~al.(2021{\natexlab{a}})Zhang, Deng, Li, Yuan, Bing, and Lam}]{zhang-etal-2021-aspect-sentiment}
Zhang, W.; Deng, Y.; Li, X.; Yuan, Y.; Bing, L.; and Lam, W. 2021{\natexlab{a}}.
\newblock Aspect Sentiment Quad Prediction as Paraphrase Generation.
\newblock In Moens, M.-F.; Huang, X.; Specia, L.; and Yih, S. W.-t., eds., \emph{Proceedings of the 2021 Conference on Empirical Methods in Natural Language Processing}, 9209--9219. Online and Punta Cana, Dominican Republic: Association for Computational Linguistics.

\bibitem[{Zhang et~al.(2023{\natexlab{a}})Zhang, Deng, Liu, Pan, and Bing}]{zhang2023sentiment}
Zhang, W.; Deng, Y.; Liu, B.; Pan, S.~J.; and Bing, L. 2023{\natexlab{a}}.
\newblock Sentiment Analysis in the Era of Large Language Models: A Reality Check.
\newblock arXiv:2305.15005.

\bibitem[{Zhang et~al.(2021{\natexlab{b}})Zhang, Li, Deng, Bing, and Lam}]{zhang-etal-2021-towards-generative}
Zhang, W.; Li, X.; Deng, Y.; Bing, L.; and Lam, W. 2021{\natexlab{b}}.
\newblock Towards Generative Aspect-Based Sentiment Analysis.
\newblock In Zong, C.; Xia, F.; Li, W.; and Navigli, R., eds., \emph{Proceedings of the 59th Annual Meeting of the Association for Computational Linguistics and the 11th International Joint Conference on Natural Language Processing (Volume 2: Short Papers)}, 504--510. Online: Association for Computational Linguistics.

\bibitem[{Zhang et~al.(2023{\natexlab{b}})Zhang, Li, Deng, Bing, and Lam}]{zhang2023survey}
Zhang, W.; Li, X.; Deng, Y.; Bing, L.; and Lam, W. 2023{\natexlab{b}}.
\newblock A Survey on Aspect-Based Sentiment Analysis: Tasks, Methods, and Challenges.
\newblock \emph{IEEE Transactions on Knowledge and Data Engineering}, 35(11): 11019--11038.

\bibitem[{Zhang et~al.(2022)Zhang, Yang, Li, Liang, Chen, Dang, Yang, and Xu}]{zhang-etal-2022-boundary}
Zhang, Y.; Yang, Y.; Li, Y.; Liang, B.; Chen, S.; Dang, Y.; Yang, M.; and Xu, R. 2022.
\newblock Boundary-Driven Table-Filling for Aspect Sentiment Triplet Extraction.
\newblock In Goldberg, Y.; Kozareva, Z.; and Zhang, Y., eds., \emph{Proceedings of the 2022 Conference on Empirical Methods in Natural Language Processing}, 6485--6498. Abu Dhabi, United Arab Emirates: Association for Computational Linguistics.

\bibitem[{Zhao et~al.(2023{\natexlab{a}})Zhao, Zhao, Lu, Wang, Tong, and Qin}]{zhao2023chatgpt}
Zhao, W.; Zhao, Y.; Lu, X.; Wang, S.; Tong, Y.; and Qin, B. 2023{\natexlab{a}}.
\newblock Is ChatGPT Equipped with Emotional Dialogue Capabilities?
\newblock arXiv:2304.09582.

\bibitem[{Zhao et~al.(2023{\natexlab{b}})Zhao, Zhou, Li, Tang, Wang, Hou, Min, Zhang, Zhang, Dong, Du, Yang, Chen, Chen, Jiang, Ren, Li, Tang, Liu, Liu, Nie, and Wen}]{zhao2023survey}
Zhao, W.~X.; Zhou, K.; Li, J.; Tang, T.; Wang, X.; Hou, Y.; Min, Y.; Zhang, B.; Zhang, J.; Dong, Z.; Du, Y.; Yang, C.; Chen, Y.; Chen, Z.; Jiang, J.; Ren, R.; Li, Y.; Tang, X.; Liu, Z.; Liu, P.; Nie, J.-Y.; and Wen, J.-R. 2023{\natexlab{b}}.
\newblock A Survey of Large Language Models.
\newblock arXiv:2303.18223.

\bibitem[{Zheng et~al.(2024)Zheng, Zhang, Zhang, Ye, Luo, and Ma}]{zheng2024llamafactory}
Zheng, Y.; Zhang, R.; Zhang, J.; Ye, Y.; Luo, Z.; and Ma, Y. 2024.
\newblock LlamaFactory: Unified Efficient Fine-Tuning of 100+ Language Models.
\newblock \emph{arXiv preprint arXiv:2403.13372}.

\bibitem[{Zhou et~al.(2024)Zhou, Wu, Song, Hu, Tian, and Xu}]{zhou2024diaasq}
Zhou, C.; Wu, Z.; Song, D.; Hu, L.; Tian, Y.; and Xu, J. 2024.
\newblock Span-Pair Interaction and Tagging for Dialogue-Level Aspect-Based Sentiment Quadruple Analysis.
\newblock In \emph{Proceedings of the ACM on Web Conference 2024}, WWW '24, 3995–4005. New York, NY, USA: Association for Computing Machinery.
\newblock ISBN 9798400701719.

\end{thebibliography}

\end{document}